\documentclass{article}

\usepackage{microtype}
\usepackage{graphicx}
\usepackage{subfigure}
\usepackage{booktabs} % for professional tables

\usepackage{amsmath,amssymb, amsthm}
\usepackage[colorlinks=true,linkcolor=blue,citecolor=blue]{hyperref}
\newtheorem{theorem}{Theorem}

\usepackage{graphicx}
\usepackage[english]{babel}
\usepackage{comment}
\usepackage[T1]{fontenc}

\usepackage[round,longnamesfirst]{natbib}

\newtheorem{innercustomthm}{Theorem}
\newenvironment{customthm}[1]
  {\renewcommand\theinnercustomthm{#1}\innercustomthm}
  {\endinnercustomthm}

\newcommand{\btheta}{\mathbf{\theta}}

\newcommand{\cO}{\mathcal{O}}

\newcommand{\eps}{\epsilon}

\def\<{\langle}
\def\>{\rangle}

\def\shownotes{1}  %set 1 to show author notes
\ifnum\shownotes=1
\newcommand{\authnote}[2]{$\ll$\textsf{\footnotesize #1 notes: #2}$\gg$}
\else
\newcommand{\authnote}[2]{}
\fi

\usepackage{hyperref}

\usepackage[accepted]{icml2020}

\icmltitlerunning{On Second-Order Group Influence Functions for Black-Box Predictions}

\begin{document}

\twocolumn[
\icmltitle{On Second-Order Group Influence Functions for Black-Box Predictions}




\begin{icmlauthorlist}
\icmlauthor{Samyadeep Basu}{to}
\icmlauthor{Xuchen You}{to}
\icmlauthor{Soheil Feizi}{to}

\end{icmlauthorlist}

\icmlaffiliation{to}{Department of Computer Science, University of Maryland-College Park}

\icmlcorrespondingauthor{Samyadeep Basu}{sbasu12@cs.umd.edu}

\icmlkeywords{Machine Learning, ICML}

\vskip 0.3in
]

\printAffiliationsAndNotice{\icmlEqualContribution}

\begin{abstract}
With the rapid adoption of machine learning systems in sensitive applications, there is an increasing need to make black-box models explainable. Often we want to identify an {\it influential} group of training samples in a particular test prediction for a given machine learning model. Existing influence functions tackle this problem by using first-order approximations of the effect of removing a sample from the training set on model parameters. To compute the influence of a group of training samples (rather than an individual point) in model predictions, the change in optimal model parameters after removing that group from the training set can be large. Thus, in such cases, the first-order approximation can be loose. In this paper, we address this issue and propose second-order influence functions for identifying influential groups in test-time predictions. For linear models, across different sizes and types of groups, we show that using the proposed second-order influence function improves the correlation between the computed influence values and the ground truth ones. We also show that second-order influence functions could be used with optimization techniques to improve the selection of the most influential group for a test-sample.  
\end{abstract}
\vspace{-0.74cm}
\section{Introduction}
\label{submission}
Recently, there has been a rapid and significant success in applying machine learning methods to a wide range of applications including vision \citep{Szeliski:2010:CVA:1941882}, natural language processing \citep{Sebastiani:2002:MLA:505282.505283},  medicine \citep{medical_imaging}, finance \citep{finance}, etc. In sensitive applications such as medicine, we would like to explain test-time model predictions to humans. An important question is : \textit{why the model makes a certain prediction for a particular test sample.} 
%One way to answer this question is through the input features of the test sample for the particular prediction. This question is often answered through the lens of saliency maps like \citep{saliency1, saliency2, saliency3,higher_order_loss}.
One way to address this is to trace back model predictions to its training data. More specifically, one can ask which training samples were the most influential ones for a given test prediction. 

Influence functions \citep{cook_influence} from robust statistics measure the dependency of optimal model parameters on training samples. Previously \citep{influence1} used first-order approximations of influence functions to estimate how much model parameters would change if a training point was up-weighted by an infinitesimal amount. Such an approximation can be used to identify most influential training samples in a test prediction. Moreover, this approximation is similar to the leave-one-out re-training, thus the first-order influence function proposed in \citep{influence1} bypasses the expensive process of repeated re-training the model to find influential training samples in a test-time prediction. 

In some applications, one may want to understand how model parameters would change when large groups of training samples are removed from the training set. This could be useful to identify groups of training data which drive the decision for a particular test prediction. 
As shown in \citep{influence2}, finding influential groups can be useful in real-world applications such as diagnosing batch effects \citep{sc-Batch}, apportioning credit between different data sources \citep{credit}, understanding effects of different demographic groups \citep{demograph} or in a multi-party learning setting \citep{multiparty}. \citep{influence2} approximates the group influence by sum of first-order individual influences over training samples in the considered group. However, removal of a large group from training can lead to a large perturbation to model parameters. Therefore, influence functions based on first-order approximations may not be accurate in this setup. Moreover, approximating the group influence by adding individual sample influences ignores possible cross correlations that may exist among samples in the group.  

In this paper, we relax the first-order approximations of current influence functions and study how second-order approximations can be used to capture model changes when a potentially large group of training samples is up-weighted. Considering a training set $\mathcal{S}$ and a group $\mathcal{U} \subset \mathcal{S}$, existing first-order approximations of the group influence function \citep{influence2} can be written as the sum of first-order influences of individual points. That is,
\begin{equation}
    \mathcal{I}^{(1)}(\mathcal{U}) = \sum_{i=1}^{|\mathcal{U}|} \mathcal{I}^{(1)}_{i} \nonumber
\end{equation}
where $\mathcal{I}^{(1)}(\mathcal{U})$ is the first-order group influence function and $\mathcal{I}^{(1)}_{i}$ is the first-order influence for the $i^{th}$ sample in $\mathcal{U}$. On the other hand, our proposed second-order group influence function has the following form:
\begin{equation}
    \mathcal{I}^{(2)}(\mathcal{U}) = \mathcal{I}^{(1)}(\mathcal{U}) + \mathcal{I}^{'}(\mathcal{U}) \nonumber
\end{equation}
where $\mathcal{I}^{'}(\mathcal{U})$ captures informative cross-dependencies among samples in the group and is a function of gradient vectors and the Hessian matrix evaluated at the optimal model parameters. We present a more precise statement of this result in Theorem \ref{main_lemma}. We note that the proposed second-order influence function can be computed efficiently even for large models. We discuss its computational complexity in Section \ref{complexity}.    

Our analysis shows that the proposed second-order influence function captures model changes efficiently even when the size of the groups are relatively large or the changes to the model parameters are significant as in the case of groups with similar properties. For example, in an MNIST classification problem using logistic regression, when $50\%$ of the training samples are removed, the correlation between the ground truth estimate and second-order influence values improves by over $55 \%$ when compared to the existing first-order influence values. We note that higher-order influence functions have been used in statistics \citep{highorder} for point and interval estimates of non-linear functionals in parameteric, semi-parametric and non-parametric models. However, to the best of our knowledge, this is the first time, higher-order influence functions are used for the interpretability task in the machine learning community. 

Similar to \citep{influence1} and \citep{influence2}, our main results for the second-order influence functions hold for linear prediction models where the underlying optimization is convex. However, we also additionally explore effectiveness of both first-order and second-order group influence functions in the case of deep neural networks. We observe that none of the methods provide good estimates of the ground-truth influence across different groups \footnote{Note that experiments of \citep{influence1} focus only on the most influential individual training samples.}. %We observe that the influence estimates in case of deep neural networks are very low when compared to the ground truth influences. We take the eigen-decomposition of the Hessian matrix in the first-order influence function and empirically show that the contribution to the influence score from each component of the eigen-decomposition is low when compared to linear models.
In summary, we make the following contributions: 
\begin{itemize}
    \item We propose second-order group influence functions that consider cross dependencies among the samples in the considered group. 
    \item Through several experiments over linear models, across different sizes and types of groups, we show that the second-order influence estimates have higher correlations with the ground truth when compared to the first-order ones, especially when the changes to the underlying model is relatively large. 
    \item We also show that our proposed second-order group influence function can be used to improve the selection of the most influential training group.
    %We specifically show that second-order group influence functions are particularly effective when the changes to the underlying model are relatively large. 
    %\item In the case of deep neural networks, we show that neither first nor the second order influence functions provide good estimates of the ground truth in case of groups. %We show that the contribution to the influence score from each eigenvector of the Hessian matrix is smaller compared to that of linear models. 
    %We explain this phenomenon by analyzing the inner products between sample gradients and eigenvectors of the Hessian matrix.
\end{itemize}

\section{Related Works}
Influence functions, a classical technique from robust statistics introduced by \citep{cook_influence, cook_inf_2} were first used in the machine learning community for interpretability by \citep{influence1} to approximate the effect of upweighting a training point on the model parameters and test-loss for a particular test sample.
%More recently \citep{pmlr_influence} focused on the behaviour of influence functions on self-loss.
In the past few years, there has been an increase in the applications of influence functions for a variety of machine learning tasks. \citep{Schulam2019CanYT} used influence functions to produce confidence intervals for a prediction and to audit the reliability of predictions. \citep{model_fairness} used influence functions to approximate the gradient in order to recover a counterfactual distribution and increase model fairness, while \citep{influence_word_embeddings} used influence functions to understand the origins of bias in word-embeddings. \citep{koh2019stronger} crafted stronger data poisoning attacks using influence functions. Influence functions can also be used to detect extrapolation \citep{extrapolation} in certain specific cases, validate causal inference models \citep{causal} and identify influential pre-training points \citep{multistage_influence}. Infinitesimal jackknife or the delta method are ideas closely related to influence functions for linear approximations of leave-one-out cross validation \citep{infinitesimal_jackknife, efron_jackknife}. Recently a higher-order instance \citep{Giordano2019AHS} of infinitesimal jackknife \citep{infinitesimal_jackknife} was used to approximate cross-validation procedures. While their setting corresponding to approximations of leave-$k$-out re-training is relatively similar to our paper, our higher-order terms preserve the empirical weight distribution of the training data in the ERM and are derived from influence functions, while in \citep{Giordano2019AHS} instances of infinitesimal jackknife is used. These differences lead to our higher-order terms being marginally different than the one proposed in \cite{Giordano2019AHS}. Our proposed second-order approximation for group influence function is additionally backed by a thorough empirical study across different settings in the case of linear models which has not yet been explored in prior works.

\section{Background}
We consider the classical supervised learning problem setup, where the task is to learn a function $h$ (also called the hypothesis) mapping from the input space $\mathcal{X}$ to an output space $\mathcal{Y}$. We denote the input-output pair as $\{x,y\}$. We assume that our learning algorithm is given training examples $\mathcal{S}:= \{ z_{i} = (x_{i}, y_{i}) \}_{i=1}^{m}$ drawn i.i.d from some unknown distribution $\mathcal{P}$. Let $\Theta $ be the space of the parameters of considered hypothesis class. The goal is to select model parameters $\btheta$ to minimize the empirical risk as follows:
\begin{equation}
\min_{\btheta\in \Theta }~~ L_{\emptyset}(\btheta) := \frac{1}{|\mathcal{S}|} \sum_{z \in \mathcal{S}}^{} \ell(h_{\btheta}(z)), 
\end{equation}
where $|\mathcal{S}|=m$, denotes the cardinality of the training set, the subscript $\emptyset$ indicates that the whole set $\mathcal{S}$ is used in training and $\ell$ is the associated loss function. We refer to the optimal parameters computed by the above optimization as $\btheta^{*}$.

Let $\nabla_{\btheta} L_{\emptyset}(\btheta)$ and $H_{ \btheta^{*}} = \nabla_{\btheta}^{2}L_{\emptyset}(\btheta)$ be the gradient and the Hessian of the loss function, respectively.

First, we discuss the case where we want to compute the effect of an {\it individual} training sample $z$ on optimal model parameters as well as the test predictions made by the model. The effect or influence of a training sample on the model parameters could characterized by removing that particular training sample and retraining the model again as follows: 
\begin{equation}
    \btheta^{*}_{\{z\}} = \arg \min_{\btheta \in  \Theta}~~ L_{\{z\}}(\theta) = \frac{1}{|\mathcal{S}|-1}\sum_{z_{i} \neq z} \ell(h_{\btheta}(z_{i}))
\end{equation}
Then, we can compute the change in model parameters as $\Delta \btheta = \btheta^{*}_{\{z\}} - \btheta^{*}$, due to removal of a training point $z$. However, re-training the model for every such training sample is expensive when $|\mathcal{S}|$ is large. Influence functions based on first-order approximations introduced by \citep{cook_influence, cook_inf_2} was used by \citep{influence1} to approximate this change. Up-weighting a training point $z$ by an infinitesimal amount $\epsilon$ leads to a new optimal model parameters, $\btheta^{\epsilon}_{\{z\}}$, obtained by solving the following optimization problem: 
\begin{equation}
   \btheta_{\{z\}}^{\epsilon} = \arg \min_{\btheta \in  \Theta}~~ \frac{1}{|\mathcal{S}|} \sum_{z \in \mathcal{S}}^{} \ell(h_{\btheta}(z_{i})) + \epsilon \ell(h_{\btheta}(z)) 
\end{equation}
%Previously \citep{influence1} studied the change in model parameters due to removing a point from the training example set. Considering a set of training examples $\{ z_{i} = (x_{i}, y_{i}) \}_{i=1}^{n}$ from which one training example $z$ is removed, the model parameters change by $\hat{\theta}_{-z} - \hat{\theta} $. More specifically $\hat{\theta}_{-z}$ is obtained by solving the following optimization formulation: %
Removing a point $z$ is similar to up-weighting its corresponding weight by $\epsilon = -\frac{1}{|\mathcal{S}|}$. The main idea used by \citep{influence1} is to approximate $\btheta^{*}_{\{z\}}$ by minimizing the first-order Taylor series approximation around $\btheta^{*}$.
Following the classical result by \citep{cook_influence}, the change in the model parameters $\btheta^{*}$ on up-weighting $z$ can be approximated by the influence function \citep{influence1} denoted by $\mathcal{I}$: 
\begin{equation}
    \mathcal{I}(z)= \frac{d \btheta^{\epsilon}_{\{ z\}}}{d \epsilon} |_{\epsilon = 0} = - H_{\btheta^{*}}^{-1} \nabla_{\btheta} \ell\left(h_{\btheta^{*}}(z)\right)
\end{equation}
A detailed proof can be found in \citep{influence1}. Using the given formulation, we can track the change with respect to any function of $\btheta^{*}$. The change in the test loss for a particular test point $z_{t}$ when a training point $z$ is up-weighted can be approximated as a closed form expression: 
\begin{align}
    \mathcal{I}(z, z_{t})=-\nabla_{\theta} \ell(h_{\btheta^{*}}(z_{t}))^{T}H_{\btheta^{*}}^{-1} \nabla_{\theta}\ell(h_{\btheta^{*}}(z))
\end{align}
This result is based on the assumption \citep{influence1} that the loss function $L(\btheta)$ is strictly convex in the model parameters $\btheta$ and the Hessian $H_{\btheta^{*}}$ is therefore positive-definite. This approximation is very similar to forming a quadratic approximation around the optimal parameters $\btheta^{*}$ and taking a single Newton step. However explicitly computing $H_{\btheta^{*}}$ and it's inverse $H_{\btheta^{*}}^{-1}$ is not required. Using the Hessian-vector product rule \citep{Pearlmutter} influence functions can be computed efficiently.

\section{Group Influence Function}
\label{def: group}
Our goal in this section is to understand how the model parameters would change if a particular group of samples was up-weighted from the training set. However, up-weighting a group can lead to large perturbations to the training data distribution and therefore model parameters, which does not follow the small perturbation assumption of the first-order influence functions. In this section, we extend influence functions using second-order approximations to better capture changes in model parameters due to up-weighting a group of training samples. In Section \ref{group: selection}, we show that our proposed second-order group influence function can be used in conjunction with optimization techniques to select the most influential training groups in a test prediction.  

The empirical risk minimization (ERM) when we remove $\mathcal{U}$ samples from training can be written as: 
\begin{equation}
    L_{\mathcal{U}}(\btheta) = \frac{1}{|\mathcal{S}|-|\mathcal{U}|}\sum_{z\in{\mathcal{S}\setminus\mathcal{U}}} \ell (h_{\btheta}(z))
\end{equation}

To approximate how optimal solution of this optimization is related to $\theta^*$, we study the effect of {\it up-weighting} a group of training samples on model parameters. Note that in this case, updated weights should still be a valid distribution, i.e. if a group of training samples has been up-weighted, the rest of samples should be down-weighted to preserve the sum to one constraint of weights in the ERM formulation. In the individual influence function case (when the size of the group is one), up-weighting a sample by $\epsilon$ leads to down-weighting other samples by $\epsilon/(m-1)$ whose effect can be neglected similar to the formulation of \citep{influence1}. 
In our formulation for the group influence function, we assume that the weights of samples in the set $\mathcal{U}$ has been up-weighted all by $\eps$ and use $p = \frac{|\mathcal{U}|}{|\mathcal{S}|}$ to denote the fraction of  up-weighted training samples. This leads to a down-weighting of the rest of training samples by $\tilde{\eps}= \frac{|\mathcal{U}|}{|\mathcal{S}| - |\mathcal{U}|}\epsilon$, to preserve the empirical weight distributioxn of the training data. This is also important in order to have a fair comparison with the ground-truth leave-out-retraining estimates. Therefore, the resulting ERM can be written as: 

$$
\btheta^{\eps}_{\mathcal{U}}=\arg\min_{\btheta} L^\epsilon_{\mathcal{U}}(\btheta)    
$$
where
\begin{align}
    L^\epsilon_{\mathcal{U}}(\btheta) =\frac{1}{|\mathcal{S}|} \Big(&\sum_{z\in{\mathcal{S}\setminus\mathcal{U}}}(1-\Tilde{\epsilon}) \ell (h_{\btheta}(z))\\
    &+ \sum_{z\in\mathcal{U}}(1+\epsilon) \ell (h_{\btheta}(z))\Big).\nonumber
\end{align}
Or equivalently
%\begin{align}
%L^\epsilon_{\mathcal{U}}(\btheta) = %L_\emptyset(\btheta) + 
%\frac{1}{|\mathcal{S}|}
%\Big(\sum_{z\in{\mathcal{S}\setminus\mathcal{U}}}-\Tilde{\epsilon}\ell(h_{\btheta}(z))\\
%+ \sum_{z\in\mathcal{U}}\epsilon \ell(h_{\btheta}(z))\Big)\nonumber
%\end{align}
%where $\Tilde{\epsilon} = \frac{|\mathcal{U}|}{|\mathcal{S}|-|\mathcal{U}|} \epsilon$. 
In the above formulation, if $\epsilon = 0$ we get the original loss function $L_{\emptyset}(\btheta)$ (where none of the training samples are removed) and if $\epsilon = -1$, we get the loss function $L_{
\mathcal{U}}(\btheta)$ (where samples %$\cU$ 
are removed from training). 

Let $\btheta^{\epsilon}_{\mathcal{U}}$ denote the optimal parameters for $L_{\mathcal{U}}^{\epsilon}$ minimization. %(the subscript $\mathcal{U}$ will be omitted when the removed set $\mathcal{U}$ is clear in the context ) and $\btheta^{*}$ denote the optimal parameters for $L_{\mathcal{U}}^{0}(\btheta)$.\\ \\%{\bf S: confusing notation, wan't $\btheta^{(0)}$ for the full ERM case?}: 
Essentially we are concerned about the change in the model parameters (i.e. $\Delta \btheta = \btheta^{\epsilon}_{\mathcal{U}} - \btheta^{*}$) when each training sample in a group of size $|\mathcal{U}|$ is upweighted by a factor of $\epsilon$. The key step of the derivation is to expand  $\btheta^{\epsilon}_{\mathcal{U}}$ around $\btheta^{*}$ (the minimizer of $L_{\mathcal{U}}^{0}(\btheta)$, or $L_{\emptyset}(\btheta)$) with respect to the order of $\epsilon$, the upweighting parameter.  In order to do that, we use the perturbation theory \citep{perturbation} to expand $\btheta^{\epsilon}_{\mathcal{U}}$ around $\btheta^{*}$. 

Frequently used in quantum mechanics and also in other areas of physics such as particle physics, condensed matter and atomic physics, perturbation theory finds approximate solution to a problem ($\btheta^{\epsilon}_{\mathcal{U}}$) by starting from the exact solution of a closely related and simpler problem ($\btheta^{*}$). As $\epsilon$ gets smaller and smaller, these higher order terms become less significant. However, for large model perturbations (such as the case of group influence functions), using higher-order terms can reduce approximation errors significantly. The following perturbation series forms the core of our derivation for second-order influence functions:
\begin{equation}
\label{perturbation_series}
    \btheta^{\epsilon}_{\mathcal{U}} - \btheta^{*} = \cO(\epsilon) \btheta^{(1)} + \cO(\epsilon^2) \btheta^{(2)}+ \cO(\epsilon^3) \btheta^{(3)} +\cdots
\end{equation}
where $\theta^{(1)}$ characterizes the first-order (in $\eps$) perturbation vector of model parameters while $\theta^{(2)}$ is the second-order (in $\eps$) model perturbation vector. We hide the dependencies of these perturbation vectors to constants (such as $|U|$) with the $\cO(.)$ notation.

In the case of computing influence of individual points, as shown by \citep{influence1}, the scaling of $\theta^{(1)}$ is in the order of $1/|\mathcal{S}|$ while the scaling of the second-order coefficient is $1/|\mathcal{S}|^{2}$ which is very small when $\mathcal{S}$ is large. Thus, in this case, the second-order term can be ignored. In the case of computing the group influence, the second-order coefficient is in the order of $|\mathcal{U}|^{2}/|\mathcal{S}|^{2}$, which can be large when the size of $\mathcal{U}$ is large. Thus, in our definition of the group influence function, both $\btheta^{(1)}$ and $\btheta^{(2)}$ are taken into account. 

The first-order group influence function (denoted by $\mathcal{I}^{(1)}$) when all the samples in a group $\mathcal{U}$ are up-weighted by $\eps$ can be defined as:
\begin{equation}
    \mathcal{I}^{(1)}(\mathcal{U}) = \frac{\partial \btheta^{\eps}_{\mathcal{U}}}{\partial \eps} |_{\eps = 0} \nonumber 
\end{equation}
\begin{equation}
    = \frac{\partial (\btheta^{*} + \cO(\epsilon)\btheta^{(1)} + \cO(\eps^{2})\btheta^{(2)}) }{\partial \eps}|_{\eps=0} = \btheta^{(1)} \nonumber 
\end{equation} \\
To capture the dependency of the terms in $\cO(\eps^{2})$, on the group influence function, we define $\mathcal{I}^{'}$ as follows:
\begin{equation}
    \mathcal{I}^{'}(\mathcal{U}) = \frac{\partial^{2} \btheta^{\eps}_{\mathcal{U}}}{\partial \eps^{2}} |_{\eps=0} \nonumber 
\end{equation}
\begin{equation}
= \frac{\partial^{2} (\btheta^{*} + \cO(\epsilon)\btheta^{(1)} + \cO(\eps^{2})\btheta^{(2)}) }{\partial \eps^{2}}|_{\eps = 0} = \btheta^{(2)} \nonumber
\end{equation} \\
Although one can consider even higher-order terms, in this paper, we restrict our derivations up to the second-order approximations of the group influence function. We now state our main result in the following theorem:
\begin{theorem}
\label{main_lemma}
If the third-derivative of the loss function at $\theta^*$ is sufficiently small, the second-order group influence function (denoted by $\mathcal{I}^{(2)}(\mathcal{U})$) when all samples in a group $\mathcal{U}$ are up-weighted by $\eps$ is:
\begin{equation}
    \mathcal{I}^{(2)}(\mathcal{U}) = \mathcal{I}^{(1)}(\mathcal{U}) + \mathcal{I}^{'}(\mathcal{U})
\end{equation}
where:
\begin{equation}
    \mathcal{I}^{(1)}(\mathcal{U}) = %{\bf XXXXXXXXX} \nonumber
    -\frac{1}{1-p}\frac{1}{|\mathcal{S}|} %(\nabla^2L_\emptyset(\btheta^{*}))^{-1} 
H_{\btheta^{*}}^{-1}
\sum_{z \in \mathcal{U}}\nabla \ell (h_{\btheta^{*}}(z)) \nonumber
\end{equation}
and 
\begin{align*}
   &\mathcal{I}^{'}(\mathcal{U}) =\\
   &\frac{p}{1-p}\Big( I  - (\nabla^2L_\emptyset(\btheta^{*}))^{-1}\frac{1}{|\mathcal{U}|}\sum_{z\in\mathcal{U}}\nabla^2 \ell (h_{\btheta^{*}}(z))\Big)\btheta^{(1)} \nonumber
\end{align*}
\end{theorem}
This result is based on the assumption that the third-order derivatives of the loss function at $\theta^*$ is small. For the quadratic loss, the third-order derivatives of the loss are zero. Our experiments with the cross-entropy loss function indicates that this assumption approximately holds for the classification problem as well. Below, we present a concise sketch of this result.
\begin{figure*}
  \includegraphics[width=17cm]{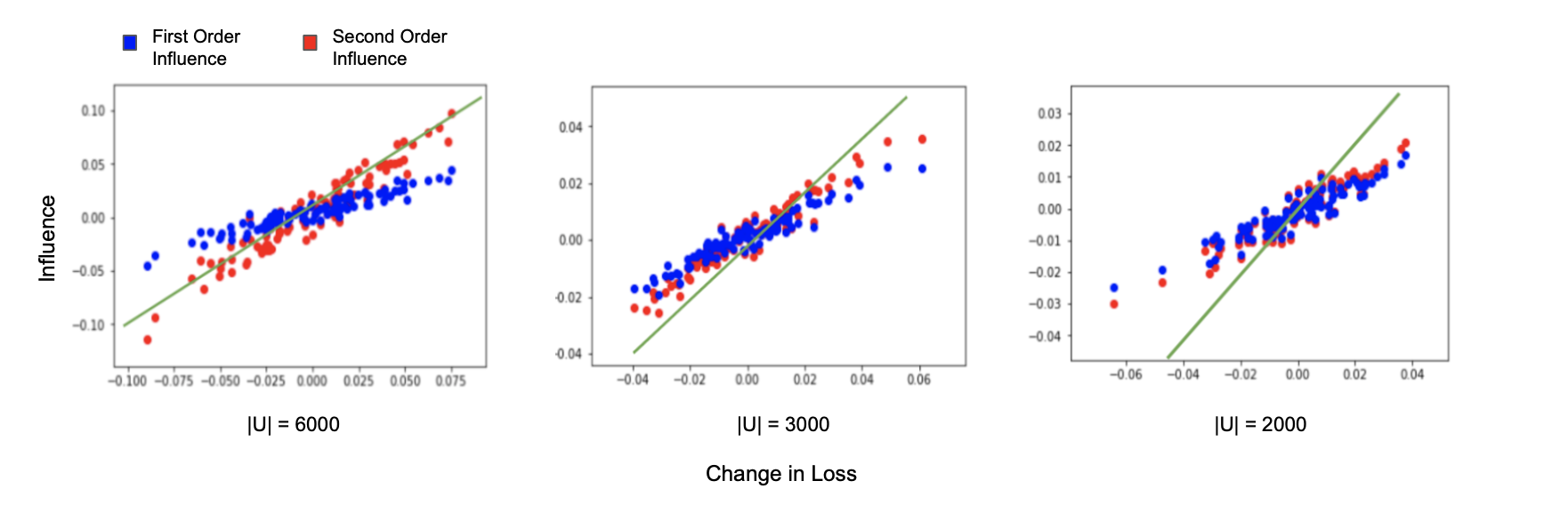}
  \caption{ \label{influence_synthetic} Comparison of first-order and second-order group influences in case of synthetic dataset with 10,000 samples using logistic regression for a mis-classified test point. Across different sizes of groups which were randomly selected, it can be observed that the second-order influence values are more correlated with the ground truth than that of the first-order ones. The green line highlights the $y=x$ line.  }
\end{figure*}
\begin{proof}[Proof Sketch]
We now derive $\btheta^{(1)}$ and $\btheta^{(2)}$ to be used in the second order group influence function $\mathcal{I}^{(2)}(\mathcal{U})$.
As $\btheta^{\epsilon}_{\mathcal{U}}$ is the optimal parameter set for the interpolated loss function $L_{\mathcal{U}}^{\epsilon}(\btheta)$, due to the first-order stationary condition, we have the following equality: 
\begin{align}
\label{main_taylor}
     0 = \nabla L_{\mathcal{U}}^\epsilon(\btheta^{\epsilon}_{\mathcal{U}}) = &\nabla L_\emptyset(\btheta^{\epsilon}_{\mathcal{U}})\\
     &+ \frac{1}{|\mathcal{S}|}(-\Tilde{\epsilon}\sum_{z \in \mathcal{S}\setminus\mathcal{U}} +\epsilon\sum_{z \in \mathcal{U}})\nabla \ell(h_{\btheta^{\epsilon}_{\mathcal{U}}}(z))\nonumber
\end{align}
The main idea is to use Taylor series for expanding $\nabla L_{\emptyset}(\btheta^{\eps}_{\mathcal{U}})$ around $\btheta^{*}$ along with the perturbation series defined in Equation (\ref{perturbation_series}) and compare the terms of the same order in $\eps$: 
\begin{equation}
    \nabla L_{\emptyset}(\btheta^{\eps}_{\mathcal{U}}) = \nabla L_{\emptyset}(\btheta^{*}) + \nabla^{2}L_{\emptyset}(\btheta^{*})(\btheta^{\eps}_{\mathcal{U}} - \btheta^{*}) +\ldots
\end{equation}
Similarly, we expand $\nabla \ell(h_{\btheta^{\eps}_{\mathcal{U}}}(z))$ around $\btheta^{*}$ using Taylor series expansion. To derive $\btheta^{(1)}$ we compared terms with the coefficient of $\cO(\eps)$ in Equation (\ref{main_taylor}) and for $\btheta^{(2)}$ we compared terms with coefficient $\cO(\eps^{2})$. Based on this, $\btheta^{(1)}$ can be written in the following way:
\begin{align}
\btheta^{(1)}  &=-\frac{1}{1-p}\frac{1}{|\mathcal{S}|} %(\nabla^2L_\emptyset(\btheta^{*}))^{-1} 
H_{\btheta^{*}}^{-1}
\sum_{z \in \mathcal{U}}\nabla \ell (h_{\btheta^{*}}(z))
\end{align}
We expand Equation(\ref{main_taylor}) and compare the terms with coefficient $\cO(\epsilon)$:
\begin{align}
    &\quad \epsilon \nabla^2L_\emptyset(\btheta^{*})\btheta^{(1)}\nonumber \\ 
    &= \frac{1}{|\mathcal{S}|} (\Tilde{\epsilon}\sum_{z \in \mathcal{S}\setminus\mathcal{U}} -\epsilon\sum_{z \in \mathcal{U}})\nabla \ell (h_{\btheta^{*}}(z)) \nonumber \\
    &= \Tilde{\epsilon}\nabla L_\emptyset(\btheta^{*}) - \frac{1}{|\mathcal{S}|}(\Tilde{\epsilon}+\epsilon)\sum_{z \in \mathcal{U}}\nabla \ell (h_{\btheta^{*}}(z)) \nonumber \\
    &= -\frac{1}{|\mathcal{S}|}(\Tilde{\epsilon}+\epsilon)\sum_{z \in \mathcal{U}}\nabla \ell(h_{\btheta^{*}}(z)) \nonumber \\
    &=-\frac{1}{|\mathcal{S}|}\frac{1}{(1 - p) }\epsilon\sum_{z \in \mathcal{U}}\nabla \ell(h_{\btheta^{*}}(z))
\end{align}

$\btheta^{(1)}$ is the first-order approximation of group influence function and can be denoted by $\mathcal{I}^{(1)}$.
Note that our first-order approximation of group influence function $\mathcal{I}^{(1)}$, is slightly different from \citep{influence2} with an additional $1-p$ in the denominator.
% This extra term arises due to the conservation of the weight distribution of the training samples, which is essential when a large group is upweighted.\\ \\
%The term $\btheta^{(2)}$ in the second order approximation can be written as:
%\begin{align}
%&\btheta^{(2)} = \frac{p}{1-p}\left( I  - (\nabla^2L_\emptyset(\btheta^{*}))^{-1}\frac{1}{|\mathcal{U}|}\sum_{z\in\mathcal{U}}\nabla^2 \ell (h_{\btheta^{*}}(z))\right)\btheta^{(1)} \nonumber \\
%&- \frac{1}{2}(\nabla^2L_\emptyset(\btheta^{*}))^{-1}\nabla^3 L_\emptyset(\btheta^{*})[\btheta^{(1)}, \btheta^{(1)},I] \nonumber \\
%\end{align}
 For $\btheta^{(2)}$ we compare the terms with coefficients of the same order of $\cO(\epsilon^2)$ in Equation (\ref{main_taylor}):
\begin{align}
 &\quad\epsilon^2 \nabla^2L_\emptyset(\btheta^{*})\btheta^{(2)} 
 + \frac{1}{2}L^{'''}_\emptyset(\btheta^{*})[\epsilon\btheta^{(1)}, \epsilon\btheta^{(1)},  I] \nonumber \\
 &+ \frac{1}{|\mathcal{S}|}(-\Tilde{\epsilon}\sum_{\mathcal{S}\setminus\mathcal{U}} +\epsilon\sum_{\mathcal{U}})\nabla^2 \ell(h_{\btheta^{*}}(z))(\epsilon\btheta^{(1)}) \nonumber \\
&= 0
\end{align}
For the $\btheta^{(2)}$ term, we ignore the third-order term $\frac{1}{2}L^{'''}_\emptyset(\btheta^{*})[\epsilon \btheta^{(1)}, \epsilon \btheta^{(1)},  I]$ due to it being small. Now we substitute the value of $\Tilde \eps$ and equate the terms with coefficient in the order of $\cO(\eps^{2})$:

\begin{align}
\label{second_order_last}
    \nabla^{2}L_{\emptyset}(\btheta^{*})\btheta^{(2)} = \frac{|\mathcal{U}|}{|\mathcal{S}| - |\mathcal{U}|} \Big(\frac{1}{|\mathcal{S}|}\sum_{z \in \mathcal{S}}\nabla^{2}\ell(h_{\btheta^{*}}(z)) \\
    - \frac{1}{|\mathcal{U}|} \sum_{z \in \mathcal{U}} \nabla^{2}\ell(h_{\btheta^{*}}(z))\Big)\btheta^{(1)} \nonumber
\end{align}
Rearranging the Equation (\ref{second_order_last}), we get the same identity as $\mathcal{I}^{'}$ in Theorem (\ref{main_lemma}).
\end{proof}
It can be observed that the additional term ($\mathcal{I}^{'}$) in our second-order approximation captures cross-dependencies among the samples in $\mathcal{U}$ through a function of gradients and Hessians of the loss function at the optimal model parameters. This makes the second-order group influence function to be more informative when training samples are correlated. In Section (\ref{experiments_total}), we empirically show that the addition of $\mathcal{I}^{'}$ improves correlation with the ground truth influence as well. 

For tracking the change in the test loss for a particular test point $z_{t}$ when a group $\mathcal{U}$ is removed, we use the chain rule to compute the influence score as follows: 
\begin{equation}
    \mathcal{I}^{(2)}(\mathcal{U},z_{t}) = \nabla \ell(h_{\btheta^{*}}(z_{t}))^{T}\left(\mathcal{I}^{(1)}(\mathcal{U}) + \mathcal{I}^{'}(\mathcal{U})\right)
\end{equation}
Our second-order approximation of group influence function consists of a first-order term that is similar to the one proposed in \citep{influence2} with an additional scaling term $1/(1-p)$. This scaling is due to the fact that our formulation preserves the empirical weight distribution constraint in ERM, which is essential when a large group is up-weighted. The second-order influence function has an additional term $\mathcal{I}^{'}$ that is directly proportional to $p$ and captures large perturbations to the model parameters more effectively. 

\section{Selection of Influential Groups}
\label{group: selection}
In this section, we explain how the second-order group influence function can be used to select the most influential group of training samples for a particular test prediction. In case of the existing first-order approximations for group influence functions, selecting the most influential group can be done greedily by ranking the training points with the highest individual influence since the group influence is the sum of influence of the individual points. However, with the second-order approximations such greedy selection is not optimal since the group influence is not additive in terms of the influence of individual points. To deal with this issue, we first decompose the second-order group influence function $\mathcal{I}^{(2)}(\mathcal{U},z_{t})$ into two terms as:
\begin{equation}
\nabla \ell(h_{\btheta^{*}}(z_{t}))^{T} \Big\{
\underbrace{\frac{1}{|\mathcal{S}|}\frac{1-2p}{(1-p)^{2}}
H_{\btheta^{*}}^{-1}
\sum_{z \in \mathcal{U}}^{}\nabla \ell(h_{\btheta^{*}}(z))}_{Term 1} +
\nonumber
\end{equation}
\begin{equation}
\underbrace{
 \frac{1}{(1-p)^{2}}\frac{1}{|\mathcal{S}|^{2}}
 \sum_{z \in \mathcal{U}}^{} H_{\btheta^{*}}^{-1} \nabla^{2} \ell(h_{\btheta^{*}}(z))
 H_{\btheta^{*}}^{-1}
 \sum_{z' \in \mathcal{U}}\nabla\ell(h_{\btheta^{*}}(z'))\Big\} }_{Term 2}
\end{equation}

where $H_{\btheta^{*}} = \nabla^2L_\emptyset(\btheta^{*}) $. While $Term1$ is additive with respect to the samples and $Term2$ has pairwise dependencies among samples. 

To simplify notation, we define the constant vector $\nabla \ell(h_{\btheta^{*}})(z_{t})^{T}H_{\btheta^{*}}^{-1}$ as $v_{1}$. Ideally for a given fixed group of size $k$, we want to find $k$ training samples amongst the total $m$ training samples which maximizes the influence for a given test point $z_{t}$. We can define this in the form of a quadratic optimization problem as follows:
\begin{align}
\label{selection_experiment}
    %\max_{w} \quad & T_{1}(w) + T_{2}(w^{2})\\ 
    \max_{w} \quad & c_{1}w^{T}a + c_{2}w^{T}Bw \\
    \textrm{s.t.} \quad & \left\lVert \nonumber w\right\rVert_{0} \leq k\\ \nonumber
\end{align}
where $B$ 
is composed of two matrices $C$ and $D$ i.e. $B = CD$. $w$ contains the weights associated with each sample in the training set. The entries of $a$ contain $v_{1}^{T} \nabla \ell(h_{\btheta^{*}}(z_{i})) ~ \forall i \in [1, m]$ and the rows of $C$ contain $v_{1}^{T} \nabla^{2} \ell(h_{\btheta^{*}}(z_{i}))H_{\btheta^{*}}^{-1} ~ \forall i \in [1, m]$. In case of $D$, the columns contain $\nabla \ell(h_{\btheta^{*}}(z_{i})) ~ \forall i \in [1, m]$. We define the constant $\frac{1}{|\mathcal{S}|}\frac{1-2p}{(1-p)^{2}}$ as $c_{1}$ and $\frac{1}{(1-p)^{2}}\frac{1}{|\mathcal{S}|^{2}}$ as $c_{2}$.
%\begin{equation}
%    T_{1}(w) = \sum_{i=1}^{|\mathcal{S}|}c_{1}v_{1}^{T}w_{i}\nabla \ell(h_{\btheta^{*}}(z_{i})) \nonumber
%\end{equation}
%\begin{equation}
% T_{2}(w^{2}) = \sum_{i,j=1}^{|\mathcal{S}|}c_{2}w_{i}w_{j}v_{1}^{T} \nabla^{2} \ell(h_{\btheta^{*}}(z_{i}))H_{\btheta^{*}}^{-1}\nabla \ell(h_{\btheta^{*}}(z_{j})) \nonumber
%\end{equation} 

This optimization can be relaxed using the $L_0-L_1$ relaxation 
as done in applications of compressed sensing \citep{Donoho:2006:CS:2263438.2272089, Candes:2005:DLP:2263433.2271950, l1approximation}. The relaxed optimization can then be solved efficiently using the projected gradient descent as denoted in \citep{projection1, Duchi:2008:EPL:1390156.1390191}. %\scomment{In our experiments, we relax the $L_{0}$ norm to $L_{1}$ norm and solve the corresponding optimization problem using projected gradient descent \cite{Duchi:2008:EPL:1390156.1390191}} \\ 

\begin{figure*}
  \includegraphics[width=17cm]{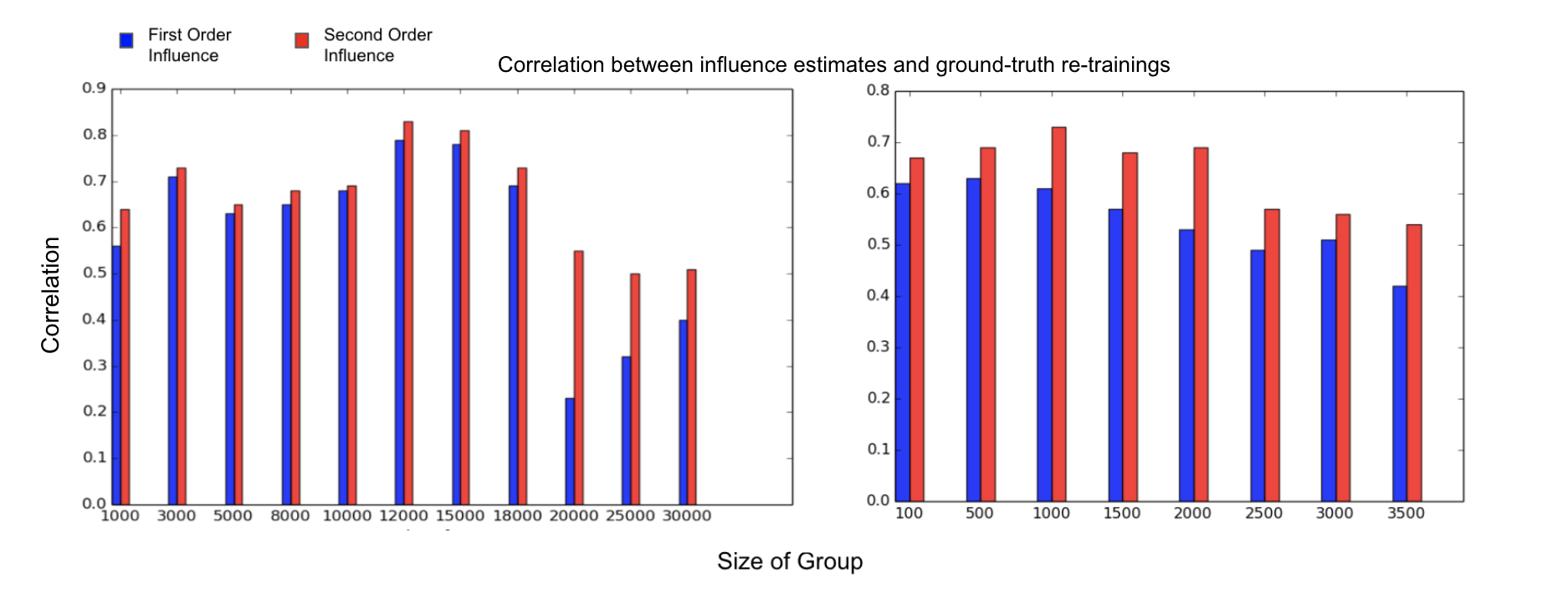}
  \caption{\label{fig_LR}Group size vs the correlation with the ground truth on MNIST for logistic regression with random groups (left panel) and {\it coherent} groups (right panel). }
\end{figure*}
\section{Computational Complexity}
\label{complexity}
For models with a relatively large number of parameters, computing the inverse of the Hessian $H_{\theta^{*}}^{-1}$ can be expensive and is of the order of $O(n^{3})$. However, computing the Hessian-vector product \citep{Pearlmutter} is relatively computationally inexpensive. In our experiments similar to \citep{influence1, influence2, multistage_influence}, we used conjugate gradients (a second-order optimization technique) \citep{conjugate_gradient} to compute the inverse Hessian-vector product which uses a Hessian-vector product in the routine thus saving the expense for inverting the Hessian directly. The proposed second-order group influence function can be computed similarly to the first-order group influence functions with only an additional step of Hessian-vector product.

\section{Experiments}
\label{experiments_total}
\subsection{Setup}
Our goal through the experiments is to observe if the second-order approximations of group influence functions improve the correlation with the ground truth estimate across different settings. We compare the computed second-order group influence score with the ground truth influence (which is computed by leave-$k$-out retraining for a group with size $k$). Our metric for evaluation is the Pearson correlation which measures how linearly the computed influence and the actual ground truth estimate are related. We perform our experiments primarily on logistic regression where the group influence function is well-defined. Additionally we also check the accuracy of first-order and second-order group influence functions in case of neural networks. 
%\subsection{Linear Models}
%Our set of experiments were conducted with logistic regression as done by \citep{influence1, influence2} for first order approximations of influence function. For both the models, the associated loss function is convex and the hessian is positive semi-definite, thus leading to relatively easier computation of inverse hessian vector product in the group influence function. 
\subsection{Datasets}
To understand the accuracy of both first-order and second-order group influence functions on linear models we use two datasets. In our first experiments, we use a synthetic dataset along with logistic regression. The synthetic dataset has 10,000 points drawn from a Gaussian distribution, consisting of 5 features and 2 classes. The details for the synthetic data can be found in the Appendix. The second set of experiments are done with the standard handwritten digits database MNIST \citep{mnist} which consists of 10 classes of different digits. For understanding how group influence functions behave in case of the neural networks we use the MNIST dataset. For each of the two datasets, we pick random groups as well {\it coherent} groups as in \citep{influence2} with sizes ranging from $1.6 \%$ to $60 \%$ of the entire training points. The computed group influence was primarily investigated for a test-point which was misclassified by the model. A detailed description of how the groups were selected in our experiments is given in the Appendix. For the optimal group selection we used a synthetic dataset consisting of 20,000 training points consisting of 5 features in the form of 4 isotropic Gaussian blobs.

% Synthetic Data plots are placed here%

%\subsection{Deep Neural Networks}
%In addition to our analysis of group influence on linear models like logistic regression, we also investigated the case of deep neural networks in this domain. The case of neural networks is important to understand in terms of interpretability, considering a lot of real-world applications nowadays are deployed with deep learning models. To the best of our knowledge, this is the first work investigating the issue of group influence functions in case of deep networks. There are two major issues associated with the case of neural networks for group influence: a) firstly the first order approximation of influence function is accurate only for a small perturbation (the error between the actual and predicted effect is small). Hence in case of group influence where the perturbation is in general high, the approximation might suffer from certain deviation. b) Secondly the loss function in case of deep neural networks is non-convex, thus breaking the assumption of positive semi-definiteness for the hessian.  Through our experiments we empirically understood how first order and second order group influence functions behave in case of neural networks. %

\subsection{Observations and Analysis}
\subsubsection{Linear Models}
For logistic regression, the general observation for the randomly selected groups was that the second-order group influence function improves the correlation with the ground truth estimates across different group sizes in both the synthetic dataset as well as MNIST. For the synthetic dataset, in Figure (\ref{influence_synthetic}), it can be observed that the approximation provided by the second-order group influence function is fairly close to the ground truth when a large fraction of the training data (60 $\%$) is removed. In such cases of large group sizes, the first-order approximation of group influence function is relatively inaccurate and far from the ground truth influence. This observation is consistent with the small perturbation assumption of first-order influence functions. However, in cases of smaller group sizes, although the second-order approximation improves over existing first-order group influence function, the gain in correlation is small. In case of MNIST, the observation was similar where the gain in correlation was significant when the size of the considered group was large. For e.g. it can seen in Figure (\ref{fig_LR}), that when more than $36 \%$ of the samples were removed, the gain in correlation is almost always more than $40 \%$.  While the improvement in correlation for larger group sizes is consistent with our theory that the second-order approximation is effective in the case of large changes to the model, the gain in correlation is non-monotonic with respect to the group sizes.
For groups of small size, selected uniformly at random, the model parameters do not change significantly and the second-order approximation improves only marginally over the existing first-order approximation. However, when a {\it coherent} group (a group having training examples from the same class) of even a relatively small size is removed, the perturbation to the model is larger (as the model parameters can change significantly in a particular direction) than if a random group is removed. In such settings, we observe that even for small group sizes, the second-order approximation consistently improves the correlation with the ground-truth significantly ({Figure (\ref{fig_LR})}). For { \it coherent} groups, across different group sizes of the MNIST dataset, we observed an improvement in correlation when the second-order approximation was used. Across different group sizes we observed that the gain in correlation is at least $15\%$. These observations (shown in Figure (\ref{fig_LR})) reinforces our theory that the second-order (or rather higher-order) approximations of influence functions are particularly effective when the perturbation or changes in the model parameters are significantly large. %However in the case of  coherent groups, although the gain in correlation by the second-order term is still non-monotonic like the random groups, the second-order approximation has a relatively monotonically decreasing behaviour (when compared to the case of random group removal) which is in accordance with the underlying theory.
The second-order approximation of the influence function could thus be used over existing first-order approximations in practical purposes such as understanding the behaviour of training groups with similar properties (e.g. demographic groups) on model predictions, without the need to actually retrain the model again.

\subsubsection{Neural Networks}
In case of neural networks, the Hessian is not positive semi-definite in general, which violates the assumptions of influence functions. Previously \citep{influence1} regularized the Hessian in the form of $H_{\btheta^{*}} + \lambda I$, and had shown that for the top few influential training points (not groups) and for a given test point, the correlation with the ground truth influence is still satisfactory, if not highly significant. However, how influence functions behave in the case of groups, is a topic not yet well explored. For MNIST, we used a regularized Hessian with a value of $\lambda = 0.01$ and conducted experiments for a relatively simple two hidden layered feed-forward network with sigmoid activations for both first-order and second-order group influence functions. The general observation was that both existing first and proposed second-order group influence functions underestimate the ground truth influence values across different group sizes, leading to a non-significant correlation. The corresponding Figure can be referred to in the Appendix. However, we observed that while the second-order influence values still suffer from the underestimation issue, they improve the correlation marginally across different group sizes. This observation was consistent in cases of both random and {\it coherent} group selections.
\begin{figure}
  \includegraphics[height = 5.75cm,width=8cm]{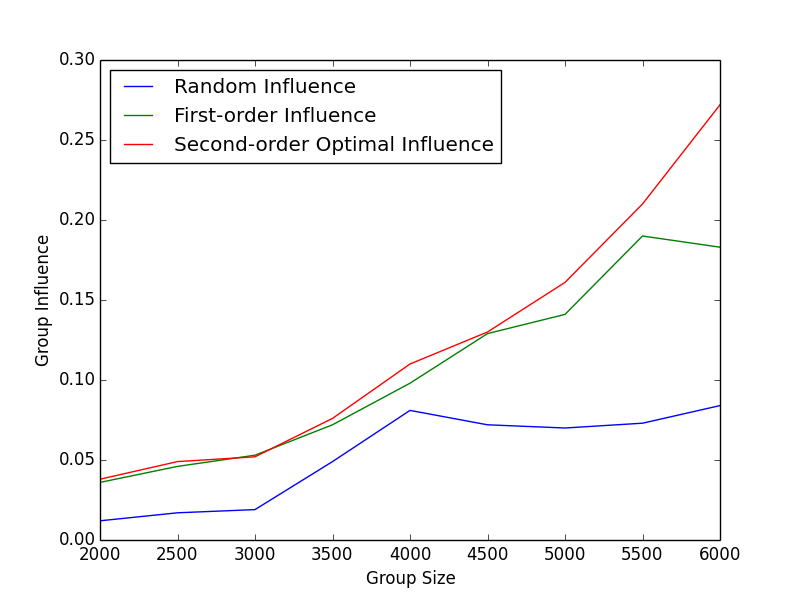}
  \caption{\label{fig_group_selection}Optimal group selection on synthetic data. }
  \vspace{-0.3cm}
\end{figure}
\subsubsection{Influential Group Selection}
In order to validate the selection of the most influential group through the second-order approximation of influence function, we performed an experiment with logistic regression (where both first-order and second-order influence function estimates are fairly close to the ground truth) on a synthetic dataset. Across different group sizes we compared the group influence (through the second-order approximation and computed with Equation (\ref{selection_experiment})) with the first-order influence computed greedily for a particular group size and the mean influence of randomly selected groups across 100 group sampling iterations. In our experiments we relaxed the $L_{0}$ norm to $L_{1}$ norm and solved the projected gradient descent step of the optimization in Equation (\ref{selection_experiment}) using \citep{Duchi:2008:EPL:1390156.1390191}. We observed that the optimal group selection procedure led to groups having relatively higher influence computed with the second-order approximation when compared to the greedy first-order influence and randomly selected groups corresponding to the different group-sizes ranging from $10\%$ to $30 \%$ of the total training samples. Specifically the optimal group influence was significantly higher than the greedy first-order group influence when the group sizes were relatively large. The selection procedure could be practically used to detect the most relevant subset of training examples which impacts a particular test-time decision through a given machine learning model when the second-order influence function is used.  
\section{Conclusion and  Future Work}
In this paper, we proposed second-order group influence functions for approximating model changes when a group from the training set is removed. Empirically, in the case of linear models, across different group sizes and types, we showed that the second-order influence has a higher correlation with ground truth values compared to the first-order ones and is more effective than existing first-order approximations. Our observation was that the second-order influence is significantly informative when the changes to the underlying model is relatively large. We showed that the proposed second-order group influence function can be practically used in conjunction with optimization techniques to select the most influential group in the training set for a particular test prediction. %However, the observed improvement in correlation by the second-order approximation is non-monotonic across different group sizes. This suggests that there is still a gap between theory and practice in the case of group influence functions which needs to be investigated further and is one interesting direction for future work. 
For non-linear models such as deep neural networks, we observed that both first-order and second-order influence functions %provide low influence scores in comparison to the ground truth values and 
lead to a non-significant correlation with the ground truth across different group sizes (although the correction values for the second-order method was marginally better). Developing accurate group influence functions for neural networks or training neural networks to have improved influence functions and also extending group influence functions to the transfer learning setting as in \cite{multistage_influence} are among directions for future work. 

% Note use of \abovespace and \belowspace to get reasonable spacing
% above and below tabular lines.

%\section*{Acknowledgements}
\nocite{langley00}
\section{Acknowledgements}
This project was supported in part by NSF CAREER AWARD 1942230, HR001119S0026-GARD-FP-052, AWS Machine Learning Research Award, a sponsorship from Capital One, and Simons Fellowship on ``Foundations of Deep Learning''.
\bibliography{example_paper}

\newpage
\section{Appendix}
\subsection{Proof for Theorem 1}

\begin{customthm}{1}
If the third-derivative of the loss function at $\theta^*$ is sufficiently small, the second-order group influence function (denoted by $\mathcal{I}^{(2)}(\mathcal{U})$) when all samples in a group $\mathcal{U}$ are up-weighted by $\eps$ is:
\begin{equation}
    \mathcal{I}^{(2)}(\mathcal{U}) = \mathcal{I}^{(1)}(\mathcal{U}) + \mathcal{I}^{'}(\mathcal{U})
\end{equation}
where:
\begin{equation}
    \mathcal{I}^{(1)}(\mathcal{U}) = %{\bf XXXXXXXXX} \nonumber
    -\frac{1}{1-p}\frac{1}{|\mathcal{S}|} %(\nabla^2L_\emptyset(\btheta^{*}))^{-1} 
H_{\btheta^{*}}^{-1}
\sum_{z \in \mathcal{U}}\nabla \ell (h_{\btheta^{*}}(z)) \nonumber
\end{equation} 
and 
\begin{align*}
   &\mathcal{I}^{'}(\mathcal{U}) =\\
   &\frac{p}{1-p}\Big( I  - (\nabla^2L_\emptyset(\btheta^{*}))^{-1}\frac{1}{|\mathcal{U}|}\sum_{z\in\mathcal{U}}\nabla^2 \ell (h_{\btheta^{*}}(z))\Big)\btheta^{(1)} \nonumber
\end{align*}
\end{customthm}
\begin{proof}
We consider the empirical risk minimization problem where the learning algorithm is given training samples $ \mathcal{S} = \{z_{i} = (x_{i},y_{i}) \}_{i=1}^{m}$ drawn i.i.d from some distribution $\mathcal{P}$. Let $\Theta $ be the space of the parameters and $h_{\btheta}$ be the hypothesis to learn. The goal is to select model parameters $\btheta$ to minimize the empirical risk as follows:
\begin{equation}
\min_{\btheta\in \Theta }~~ L_{\emptyset}(\btheta) := \frac{1}{|\mathcal{S}|} \sum_{z \in \mathcal{S}}^{} \ell(h_{\btheta}(z)), 
\end{equation}
The ERM problem with a subset $\mathcal{U} \subset \mathcal{S} $ removed from $\mathcal{S}$ is as follows:
\begin{equation}
    L_{\mathcal{U}}(\btheta) := \frac{1}{|\mathcal{S}|-|\mathcal{U}|}\sum_{z\in{\mathcal{S}\setminus\mathcal{U}}} \ell (h_{\btheta}(z))
\end{equation}
In our formulation for the group influence function, we assume that weights of samples in the set $\mathcal{U}$ has been up-weighted all by $\eps$.   This leads to a down-weighting of the remaining training samples by $\tilde{\eps}= \frac{|\mathcal{U}|}{|\mathcal{S}| - |\mathcal{U}|}\epsilon$, to conserve the empirical weight distribution of the training data. We denote $p$ as $\frac{|\mathcal{U}|}{|\mathcal{S}|}$ to denote the fraction of up-weighted training samples. Therefore, the resulting ERM optimization can be written as: 
\begin{equation}
    \btheta^{\epsilon}_{\mathcal{U}} = \arg \min_{\theta} L^{\eps}_{\mathcal{U}}(\btheta)
\end{equation}
where:
\begin{equation}
L^\epsilon_{\mathcal{U}}(\btheta) = L_\emptyset(\btheta) + 
\frac{1}{|\mathcal{S}|}
\Big(\sum_{z\in{\mathcal{S}\setminus\mathcal{U}}}-\Tilde{\epsilon}\ell(h_{\btheta}(z))+ \sum_{z\in\mathcal{U}}\epsilon \ell(h_{\btheta}(z))\Big)
\end{equation}
and $\Tilde \epsilon = \frac{|\mathcal{U}|\eps}{|\mathcal{S}| - |\mathcal{U}|}$.
We consider the stationary condition where the gradient of $L^{\eps}_{\mathcal{U}}$ is zero. More specifically: 
\begin{align}
\label{grad}
     0 = \nabla L_{\mathcal{U}}^\epsilon(\btheta^{\epsilon}_{\mathcal{U}}) = \nabla L_\emptyset(\btheta^{\epsilon}_{\mathcal{U}}) + \frac{1}{|\mathcal{S}|}\Big(-\Tilde{\epsilon}\sum_{z \in \mathcal{S}\setminus\mathcal{U}}\nabla \ell(h_{\btheta^{\epsilon}_{\mathcal{U}}}(z))\\
     +\epsilon\sum_{z \in \mathcal{U}}\nabla \ell(h_{\btheta^{\epsilon}_{\mathcal{U}}}(z))\Big) \nonumber
\end{align}
Next we expand $\nabla L_{\emptyset}(\btheta_{\mathcal{U}}^{\eps})$ around the optimal parameter $\btheta^{*}$ using Taylor's expansion and retrieve the terms with coefficients $\cO(\epsilon)$ to find $\btheta^{(1)}$:
\begin{align}
\label{8}
 0 = \nabla L_{\emptyset}(\btheta^{*}) + \nabla^{2}L_{\emptyset}(\btheta^{*})(\btheta^{\eps}_{\mathcal{U}} - \btheta^{*}) \nonumber \\ +\frac{1}{|\mathcal{S}|}\Big(-\Tilde \eps \sum_{z \in \mathcal{S}}\nabla \ell(h_{\btheta^{*}}(z))\Big)  \\
 +\frac{1}{|\mathcal{S}|}(\eps + \Tilde \eps) \sum_{z \in \mathcal{U}}\nabla \ell(h_{\btheta^{*}}(z)) \nonumber 
\end{align}
At the optimal parameters $\btheta^{*}$, $\nabla L_{\emptyset}(\btheta^{*}) = 0$ and $\btheta_{\mathcal{U}}^{\eps} - \btheta^{*} = \eps \btheta^{(1)}$, thus simplifying Equation (\ref{8}):
\begin{align}
    &\quad \epsilon \nabla^2L_\emptyset(\btheta^{*})\btheta^{(1)}\nonumber \\ 
    &= \frac{1}{|\mathcal{S}|} (\Tilde{\epsilon}\sum_{z \in \mathcal{S}\setminus\mathcal{U}} -\epsilon\sum_{z \in \mathcal{U}})\nabla \ell (h_{\btheta^{*}}(z)) \nonumber \\
    &= \Tilde{\epsilon}\nabla L_\emptyset(\btheta^{*}) - \frac{1}{|\mathcal{S}|}(\Tilde{\epsilon}+\epsilon)\sum_{z \in \mathcal{U}}\nabla \ell (h_{\btheta^{*}}(z)) \nonumber \\
    &= -\frac{1}{|\mathcal{S}|}(\Tilde{\epsilon}+\epsilon)\sum_{z \in \mathcal{U}}\nabla \ell(h_{\btheta^{*}}(z)) \\
    &=-\frac{1}{|\mathcal{S}|}\frac{|\mathcal{S}|}{(|\mathcal{S}| - |\mathcal{U}|) }\epsilon\sum_{z \in \mathcal{U}}\nabla \ell(h_{\btheta^{*}}(z)) \nonumber
\end{align}
%\begin{equation}
%    \nabla^{2}L_{\emptyset}(\btheta^{*})(\eps \btheta^{(1)}) = -\frac{1}{|\mathcal{S}|}\frac{|\mathcal{S}|}{|\mathcal{S}| - |\mathcal{U}|} \sum_{z \in \mathcal{U}}\nabla \ell(h_{\btheta^{*}}(z)) \epsilon 
%\end{equation}
Substituting $| \mathcal{U}|/ | \mathcal{S}|$ as $p$, we get the following identity: 
\begin{equation}
\label{first_ord}
    \btheta^{(1)} = - \frac{1}{|\mathcal{S}|}\frac{1}{1-p} \Big(\nabla^{2} L_{\emptyset}(\btheta^{*}) \Big)^{-1} \sum_{z \in \mathcal{U}} \nabla \ell(h_{\btheta^{*}}(z))
\end{equation}
where $\btheta^{(1)}$ is the first-order approximation of group influence function. We denote the first-order approximation as $\mathcal{I}^{(1)}$. \\ \\
Next we derive $\mathcal{I}^{'}$ by  comparing terms to the order of $\epsilon^{(2)}$ (i.e. $\cO(\eps^{2})$) in Equation (\ref{grad}) by expanding around $\btheta^{*}$:
\begin{align}
\label{r1}
    0 =  \nabla^{2}L_{\emptyset}(\btheta^{*})(\eps \btheta^{(1)} + \eps^{2}\btheta^{(2)}+ \cdots)\\
    + \frac{1}{|\mathcal{S}|}(-\Tilde \eps \sum_{z \in \mathcal{S}}\nabla \ell(h_{\btheta^{\eps}_{\mathcal{U}}}(z)) \nonumber \\
    + \frac{|\mathcal{S}|\eps}{|\mathcal{S}|-|\mathcal{U}|}\sum_{z \in \mathcal{U}}\nabla \ell(h_{\btheta^{\eps}_{\mathcal{U}}}(z)) ) \nonumber
\end{align}
where:
\begin{equation}
\label{ref: expand}
    \nabla \ell(h_{\btheta^{\eps}_{\mathcal{U}}}(z)) = \nabla \ell(h_{\btheta^{*}}(z)) + \nabla^{2}\ell(h_{\btheta^{*}}(z))(\btheta^{\eps}_{\mathcal{U}} - \btheta^{*}) + \cdots
\end{equation}
Substituting Equation (\ref{ref: expand}) in Equation (\ref{r1}), and expanding in $\cO(\eps^{2})$ we get the following identity:
\begin{align}
\label{eq10}
    \nabla^{2}L_{\emptyset}(\btheta^{*}) \epsilon^{2}\btheta^{(2)} - \frac{1}{|\mathcal{S}|}\frac{|\mathcal{U}|}{|\mathcal{S}|-|\mathcal{U}|} \sum_{z \in \mathcal{S}}\nabla^{2} \ell(h_{\btheta^{*}}(z)) \epsilon^{2} \btheta^{(1)}\\
    + \frac{1}{|\mathcal{S}|-|\mathcal{U}|} \sum_{z \in \mathcal{U}} \nabla^{2} \ell(h_{\btheta^{*}}(z))\eps^{2}\btheta^{(1)} \nonumber
\end{align}
Taking the common coefficients $|\mathcal{U}|/ (|\mathcal{S}| - |\mathcal{U}|)$ out and rearranging Equation (\ref{eq10}), we get the following identity: 
\begin{align}
\label{r3}
    \nabla^{2}L_{\emptyset}(\btheta^{*})\btheta^{(2)} =  \frac{|\mathcal{U}|}{|\mathcal{S}| - |\mathcal{U}|} \Big(\frac{1}{|\mathcal{S}|}\sum_{z \in \mathcal{S}}\nabla^{2}\ell(h_{\btheta^{*}}(z))\\
    - \frac{1}{|\mathcal{U}|} \sum_{z \in \mathcal{U}} \nabla^{2}\ell(h_{\btheta^{*}}(z))\Big)\btheta^{(1)} \nonumber
\end{align}
Now we multiply both sides of the Equation (\ref{r3}) with the Hessian inverse i.e. $\nabla^{2}L_{\emptyset}(\btheta^{*})^{-1}$ , we obtain the cross-term involving the gradients and the Hessians of the removed points in the second order influence function as follows: 
\begin{align}
\label{second_ord}
    \btheta^{(2)} = \frac{p}{1-p}\Big(I - \frac{1}{|\mathcal{U}|} \nabla^{2} \Big(L_{\emptyset}(\btheta^{*})\Big)^{-1} \sum_{z \in \mathcal{U}} \nabla^{2} \ell(h_{\btheta^{*}}(z))\Big) \btheta^{(1)}
\end{align}
where $p = |\mathcal{U}|/|\mathcal{S|}$ and we denote $\btheta^{(2)}$ as $\mathcal{I}^{'}$. We combine Equation (\ref{first_ord}) and (\ref{second_ord}), to write the second order influence function $\mathcal{I}^{(2)}(\mathcal{U})$ as:
\begin{align}
    \mathcal{I}^{(2)}(\mathcal{U}) = \mathcal{I}^{(1)}(\mathcal{U}) + \mathcal{I}^{'}(\mathcal{U})
\end{align}
\end{proof}
\subsection{Synthetic Dataset - Influence Computation}
The synthetic data is sampled from a multivariate Gaussian distribution with 5 dimensions and consists of two classes. We sample a total of 10000 points for our experiments. For the first class, the mean is kept at 0.1, while for the second class the mean is kept at 0.8. The covariance matrix in both the classes is a diagonal matrix whose entries are sampled randomly between 0 and 1.

%\begin{figure*}
  %\includegraphics[width=17cm]{main/composite_total.png}
 % \includegraphics[width=17cm]{complete_final.png}
  %\includegraphics[width = \columnwidth]{main/k_3000_LR_new.png}
  %\caption{\label{fig_LR}Group size vs the correlation with the ground truth on MNIST for Logistic Regression (left panel) and the neural network (right panel).}
%\end{figure*}

\subsection{Coherent Groups}
In case of randomly removed groups, the perturbations to the model are relatively small when compared to a case where groups having similar properties are removed. This is also the experimental setting in \cite{influence2}, where the first-order group influence function is analysed. In our experimental setup for the MNIST dataset, we remove groups of sizes (denoted by $\mathcal{\|U\|}$) ranging from 100 to 3500 from a specific class. Essentially when a group of points are removed, they are ensured to be from a similar class.  
%\begin{figure}
  %\includegraphics[width=17cm]{main/composite_total.png}
 % \includegraphics[height = 7cm,width=8cm]{coherent_new.png}
  %\includegraphics[width = \columnwidth]{main/k_3000_LR_new.png}
  %\caption{\label{fig_LR_coherent}Group size vs the correlation with the ground truth on MNIST for Logistic Regression with coherent groups.}
%\end{figure}


%%%%%%%%%%%%%%%%%%%%%%%%%%%%%%%%%%%%%%%%%%%%%%%%%%%%%%%%%%%%%%%%%%%%%%%%%%%%%%%
%%%%%%%%%%%%%%%%%%%%%%%%%%%%%%%%%%%%%%%%%%%%%%%%%%%%%%%%%%%%%%%%%%%%%%%%%%%%%%%

\subsection{Synthetic Dataset - Group Selection}
For testing our optimal group selection procedure for second-order group influence functions we generate synthetic data consisting of 20000 samples from {\fontfamily{qcr}\selectfont
sklearn.datasets.make$_{-}$blobs 
}consisting of 5 features from 4 distinct classes. The generated data is in the form of isotropic Gaussian blobs. 
\subsection{Plots for Neural Network Experiments}
\begin{figure}[h]
\centering
 \includegraphics[height = 6cm,width=8cm]{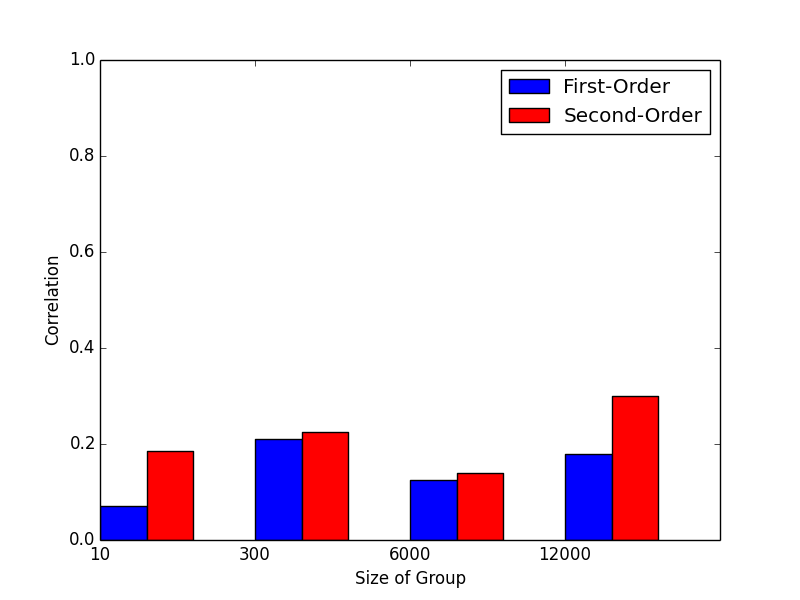}
  \caption{\label{nn_group_selection} Correlation vs group size plot for neural networks (MNIST dataset).}
  %\vspace{-1cm}
\end{figure} 
In case of neural networks, we use the existing first-order approximation of influence functions and our proposed second-order influence functions to approximate effect of leave-k-out retraining on the test-loss for a particular sample. We use the MNIST dataset in our experiments. Across different groups sizes ranging from $10\%$ to $50\%$ of the entire training data, we observe that the correlation with the ground-truth is far from significant (Figure (\ref{nn_group_selection})).  Up-weighting of a group leads to a large perturbation to the underlying model, where a local approximation via influence functions around the loss function at the optimal parameters might not lead to an accurate estimate in case of deep networks when compared to a linear model. A comprehensive investigation of the behaviour of influence functions in deep learning is a direction for future work.
\subsection{On Infinitesimal Jackknife and Influence Functions}
Infinitesimal jackknife and influence functions are methods from robust statistics to linearly approximate cross-validation procedures and attaching standard errors to point estimates \citep{efron_jackknife, infinitesimal_jackknife, cook_influence, cook_inf_2}. Previously \citep{efron_jackknife} have shown that linear approximations for cross-validation in case of infinitesimal jackknife and influence functions are relatively similar, although the technical ideas are different.  Recently \citep{Giordano2018ASA} approximated cross-validation procedures by using a linear approximation to the dependence of the cross-validation fitting procedure on the empirical weight distribution of the training data in the ERM. This procedure is similar in idea to \citep{influence1}, but differs technically where a training sample was up-weighted by an $\eps$ factor and the approximated solution to the modified ERM was found using influence functions. However in both the cases of \citep{Giordano2018ASA} and \citep{influence1, influence2} higher-order terms have been ignored. A recent work \citep{Giordano2019AHS} focuses on higher-order terms of infinitesimal jackknife for leave-k-out retraining approximations. \citep{Giordano2019AHS} specifically provide tools for extending infinitesimal jackknife\citep{Giordano2018ASA} to higher-order terms with finite sample accuracy bounds. We give a small description on how the higher-order terms in \citep{Giordano2019AHS} have been derived, in the following part: \\ \\
With a training set $\{z_{i} = (x_{i},y_{i}) \}_{i=1}^{N}$ we consider the following ERM problem:
\begin{equation}
L(\btheta, w) = \frac{1}{N} \sum_{i=1}^{N} w_{i}\ell(z_{i},\btheta)
\end{equation}
and let $\hat{\btheta}(w)$ be the solution to the following optimization problem:
\begin{equation}
    \hat{\btheta}(w) = \arg \min_{\btheta} L(\btheta, w)
\end{equation}
Let $\hat{\btheta}$ be the solution to the ERM problem with $w_{i}=1, \hspace{1em} \forall i \in [1,N]$. We denote this vector of $w$ in case of $\hat{\btheta}$ as $1_{N}$. We consider the change in the empirical weight distribution of the training data in the ERM as $\Delta{w} = w - 1_{N}$. Infinitesimal jackknife procedures compute the directional derivative of $\hat{\btheta}(w)$ in a direction denoted by $\Delta{w}$. Let $\delta_{w}^{1}\hat{\btheta}(w)$ be the first-order directional derivative of $\hat{\btheta}(w)$ in the direction $\Delta{w}$,  $\delta_{w}^{2}\hat{\btheta}(w)$ be the second-order derivative and $\delta_{w}^{k}\hat{\btheta}(w)$ be the $k^{th}$ order derivative. As a representative example, $\delta_{w}^{1}\hat{\btheta}(w)$ is defined in the following way:
\begin{equation}
    \delta_{w}^{1}\hat{\btheta}(w) = \sum_{n=1}^{N} \frac{\partial \hat{\btheta}(w)}{\partial w_{n} }|_{w} \Delta{w_{n}}
\end{equation}
and the second-order directional derivative as: 
\begin{equation}
    \delta_{w}^{2}\hat{\btheta}(w) = \sum_{n_{1}=1}^{N}\sum_{n_{2} = 1}^{N} \frac{\partial^{2} \hat{\btheta}(w)}{\partial w_{n_{1}} \partial w_{n_{2}}} |_{w} \Delta w_{n_{1}} \Delta w_{n_{2}}
\end{equation}
Infinitesimal jackknife focuses on finding the solution to $\hat{\btheta}(w)$ by a Taylor series approximation around $\hat{\btheta}$ as follows : 
\begin{equation}
    \hat{\btheta}(w) = \hat{\btheta}(1_{N}) + \sum_{i=1}^{k} \frac{1}{i!} \delta_{w}^{i} \hat{\btheta}(1_{N})
\end{equation}
\citep{Giordano2019AHS} specifically provide tools to obtain the higher-order directional derivatives i.e ($k \geq 2$) through a recursive procedure and also provide their associated finite sample accuracy bounds. However any experiment supporting how their derived higher-order infinitesimal jackknife behave in case of practical machine learning models was not explored in \citep{Giordano2019AHS}.
Note that even in influence functions $\hat{\btheta}$ is used as the known solution around which the up-weighted solution of the ERM is found. However the technical derivation of the parameters of the ERM with up-weighted training examples is different as can be observed in \citep{cook_influence, influence1, influence2}. Our work specifically focuses on higher-order terms in case of influence functions first shown in \citep{cook_influence} and subsequently by \citep{influence1, influence2}. We use a combination of perturbation series and Taylor series to compute the higher-order terms in case of influence functions. In the process we make certain practical modifications like conserving the empirical weight distribution of the training data to one, which is particularly important as shown in the main paper. Note that higher-order terms of influence functions as shown in this paper is different from the higher-order terms of infinitesimal jackknife as shown in \citep{Giordano2019AHS} due to these differences. Also we show practically how the second-order influence function is beneficial over existing first-order influence function in case of approximating leave-k-out retraining procedures for machine learning models especially for large values of k as well as for influential group selection procedure.  

\section{Efficient Computation of Second-Order Influence Function }
The second-order group influence function ($\mathcal{I}^{(2)}(\mathcal{U},z_{t})$) can be expressed as: 
\begin{equation}
\nabla \ell(h_{\btheta^{*}}(z_{t}))^{T} \Big\{
\underbrace{\frac{1}{|\mathcal{S}|}\frac{1-2p}{(1-p)^{2}}
H_{\btheta^{*}}^{-1}
\sum_{z \in \mathcal{U}}^{}\nabla \ell(h_{\btheta^{*}}(z))}_{Term 1} +
\nonumber
\end{equation}
\begin{equation}
\underbrace{
 \frac{1}{(1-p)^{2}}\frac{1}{|\mathcal{S}|^{2}}
 \sum_{z \in \mathcal{U}}^{} H_{\btheta^{*}}^{-1} \nabla^{2} \ell(h_{\btheta^{*}}(z))
 H_{\btheta^{*}}^{-1}
 \sum_{z' \in \mathcal{U}}\nabla\ell(h_{\btheta^{*}}(z'))\Big\} }_{Term 2}
\end{equation}
$Term1$ can be efficiently computed using a combination of Hessian-vector product \cite{Pearlmutter} and conjugate-gradient. $Term2$, which captures cross-dependencies amongst the samples can also be computed efficiently using three steps. First $v_{1} = H_{\btheta^{*}}^{-1}
 \sum_{z' \in \mathcal{U}}\nabla\ell(h_{\btheta^{*}}(z'))$ is computed using a combination of Hessian-vector product and conjugate gradient. In the next step, $v_{2} = \nabla^{2} \ell(h_{\btheta^{*}}(z))v_{1}$ is computed using Hessian-vector product and the output of this step is a vector. In the final step, $v_{3} = H_{\btheta^{*}}^{-1} v_{2}$ is computed using Hessian-vector product and conjugate-gradient. The second-order group influence function thus requires an extra computation of conjugate-gradient. While the second-order group influence function is a bit more expensive than the first-order one (due to an extra CG operation), it is much faster than retraining to compute the ground-truth influence. For example, the running time of first-order influence on MNIST for different groups (up to 20 $\%$ of removed data) averaged over 100 runs is 16.7 $\pm$ 2.56s while that of the second-order method is 28.2 $\pm$ 4.8s. 
 
\section{Additional Experimental Results}
\begin{figure}
 \includegraphics[height = 6cm,width=8cm]{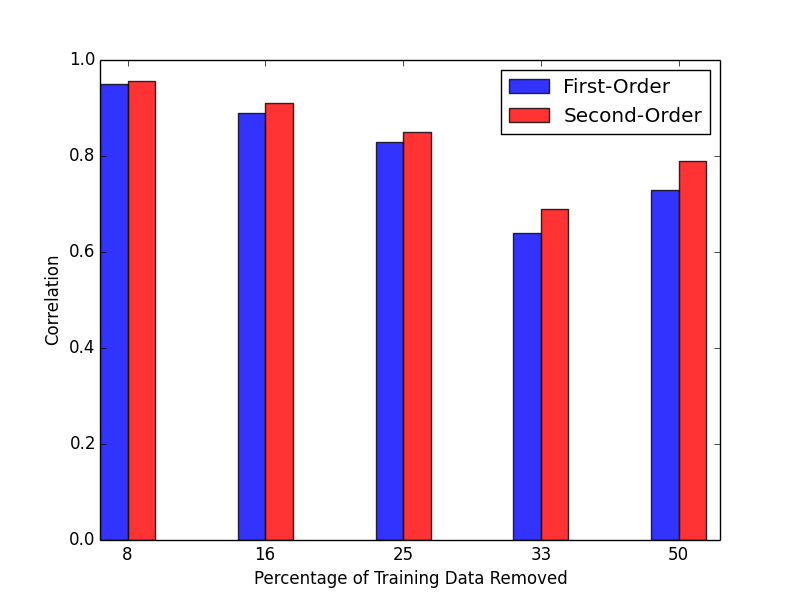}
  \caption{\label{iris_linear}Correlation vs group size plot for Iris dataset.}
\end{figure}

In this section, we provide an additional experimental result with the second-order group influence function evaluated on the Iris dataset for logistic regression. From Figure (\ref{iris_linear}), we observe that the second-order group influence estimates have a better correlation with the ground-truth across different group sizes. Specifically we notice that when more than $33 \%$ of the training data is removed, the improvement in correlation by the second-order influence function is more. In this experimental setup, the groups were removed randomly. Specifically for each group size, 200 groups were randomly removed and the experiment was repeated for 10 trials to obtain the mean correlation. 
\section{Experimental Details on MNIST}
For the experiments with the MNIST dataset, for each group size and test-point (in case of both the random and coherent groups), 500 groups were removed and the experiment is repeated for 10 trials. In our paper, we report the mean correlation amongst 10 experimental trials.  
\section{Relationship To Representer Point Theorem}
Identification of influential samples in the training set can also be done by alternative methods such as representer point theorem \cite{representer}. In particular, unlike influence functions, \cite{representer} assigns an influence score by kernel evaluations at the training samples. While both \cite{representer} and our method explain a test-prediction through the lens of the training data, the definitions of influences or importances are different in both cases. Influence functions define importance analogous to the leave-out re-training procedure while the kernel function defined in \cite{representer} evaluates influences by relying on the weighted sum of the feature similarity of the training samples in the pre-activation layer of a deep network. Investigation of the exact relationship between kernel based methods like representer theorem and influence functions is a direction for future work.

\end{document}